\documentclass[runningheads]{llncs}
\usepackage{graphicx}
\usepackage{comment}
\usepackage{amsmath,amssymb,bm,stmaryrd} 
\usepackage{booktabs} 

\usepackage[outercaption]{sidecap}  
\usepackage{color}
\usepackage{multirow}
\usepackage[normalem]{ulem}
\usepackage{enumitem}
\usepackage[width=122mm,left=12mm,paperwidth=146mm,height=193mm,top=12mm,paperheight=217mm]{geometry}
\newcommand{\etal}{\textit{et al}.}
\newcommand{\ie}{\textit{i}.\textit{e}.}
\newcommand{\eg}{\textit{e}.\textit{g}.}
\newcommand*{\affmark}[1][*]{\textsuperscript{#1}}
\usepackage[para,online,flushleft]{threeparttable}

\newcommand{\veryshortarrow}[1][3pt]{\mathrel{%
		\hbox{\rule[\dimexpr\fontdimen22\textfont2-.2pt\relax]{#1}{.4pt}}%
		\mkern-4mu\hbox{\usefont{U}{lasy}{m}{n}\symbol{41}}}}

\definecolor{azure(colorwheel)}{rgb}{0.0, 0.5, 1.0}

\begin{document}
	\pagestyle{headings}
	\mainmatter
	\def\ECCVSubNumber{4}  
	
	\title{Cross-domain Self-supervised Learning for Domain Adaptation with Few Source Labels
	} 
	

	\titlerunning{CDS: Cross-Domain Self-supervised Learning}
	%
	\author{Donghyun Kim\inst{1} \and
	Kuniaki Saito\inst{1}\and  
	Tae-Hyun Oh\inst{2} \and\\
	Bryan A. Plummer\inst{1} \and
	Stan Sclaroff\inst{1} \and 
	Kate Saenko\inst{1} }
	\authorrunning{Kim \etal}
	%
	\institute{\affmark[1]Dept. CS, Boston University, \affmark[2]Dept. EE, POSTECH\\
	\email{\{{donhk, keisaito, bplum, sclaroff, saenko\}@bu.edu}},
	\email{{taehyun@postech.ac.kr}}
	}
	\maketitle
	
	\begin{abstract}
	Existing unsupervised domain adaptation methods aim to transfer knowledge from a label-rich source domain to an unlabeled target domain. However, obtaining labels for some source domains may be very expensive, making complete labeling as used in prior work impractical. In this work, we investigate a new domain adaptation scenario with sparsely labeled source data,
		where only a few examples in the source domain have been labeled,
		while the target domain is unlabeled. We show that when labeled source examples are limited, existing methods often fail to learn discriminative features applicable for both source and target domains. We propose a novel Cross-Domain Self-supervised (CDS) learning approach for domain adaptation, which learns features that are not only domain-invariant but also class-discriminative.
		Our self-supervised learning method captures apparent visual similarity with \textit{in-domain self-supervision} in a domain adaptive manner and performs cross-domain feature matching with \textit{across-domain self-supervision}. In extensive experiments with three standard benchmark datasets, our method significantly boosts performance of target accuracy in the new target domain with few source labels and is even helpful on classical domain adaptation scenarios.
		
		
	\end{abstract}

	\section{Introduction}

\begin{figure}[t]
	\centering
	\includegraphics[width=\linewidth]{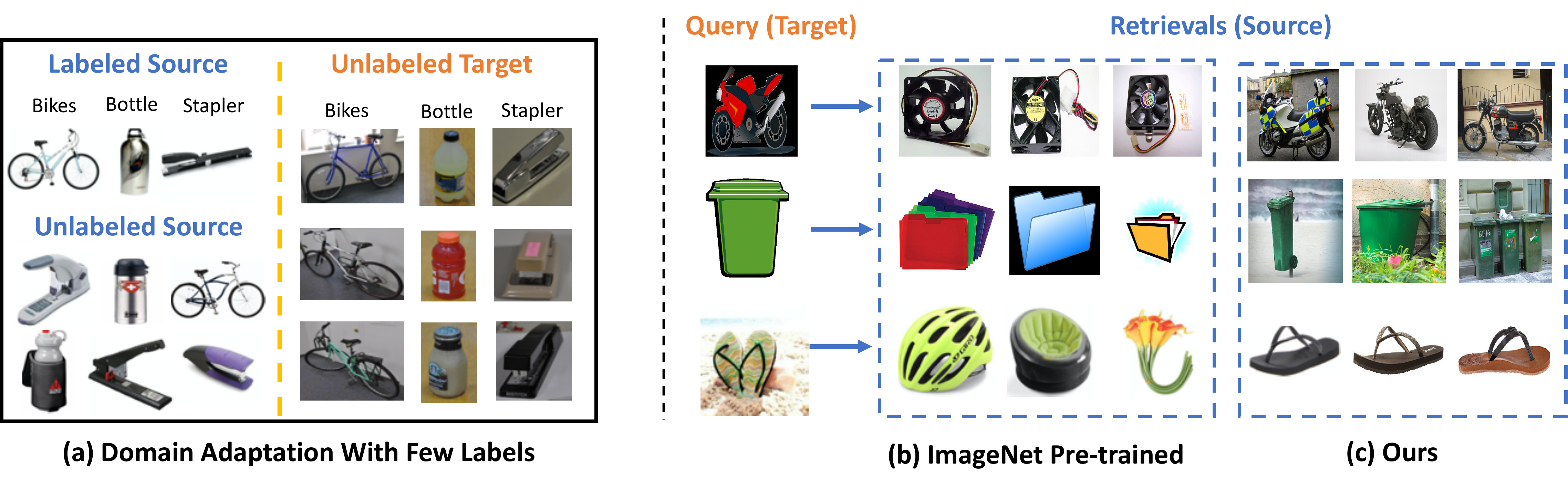}\vspace{-3mm}
	\caption{\small
	(a) An illustration of our proposed domain adaptation scenario with only a few source labels. Learning category-discriminative features on the source domain becomes more challenging, which leads to poor category knowledge transfer to the target domain. We propose a new method to learn better representations using cross-domain self-supervision. (b, c): Retrieval of the closest cross-domain neighbors using standard ImageNet-pretrained features (b) and features produced by our method (c). ImageNet-pretrained features fail to retrieve correct categories and instead embed two different classes nearby based on cues such as black and red colors (top row). In contrast, our method extracts more semantically meaningfull features that are also domain-invariant. 
	}
    \label{fig:fig1}
\end{figure}

Deep models often fail to generalize to new domains due to the \textit{domain gap} between 
data distributions at training and testing phases.
Recent unsupervised domain adaptation methods tackle this challenge by transferring knowledge from a label-rich source domain to an unlabeled new target domain (see Fig.~\ref{fig:fig1}-(a)). 
However, in practice, completing large‐scale annotation in the source domain itself is often a challenging task in itself due to the 
high cost or difficulty of annotation, \eg,~medical images should be annotated by domain experts~\cite{perone2019unsupervised}. Thus, in practical machine learning workflows, it is untenable to assume that fully annotated massive datasets are always readily available.

In this paper, we explore a new domain adaptation scenario where only a small number of examples in the source domain is annotated while the other examples remain unlabeled, \ie, sparsely labeled source data. Figure~\ref{fig:fig1}-(a) shows our proposed task, \ie, unsupervised domain adaptation with few source labels. Since we do not assume we have a fully annotated source domain dataset, our new task is more practical and challenging than the conventional unsupervised domain adaptation setup (\eg,~\cite{ganin2017domain,long2018conditional,saito2019semi,zou2019confidence}). 

State-of-the-art unsupervised domain adaptation methods (\eg,~\cite{ganin2017domain,long2018conditional,saito2019semi,zou2019confidence}) leverage a large amount of source supervision and knowledge obtained from large-scale datasets such as ImageNet~\cite{russakovsky2015imagenet}. 
They train the model on the source domain using full supervision, often starting from an ImageNet-pretrained initialization.
Then, to learn a class-discriminative representation for the target domain, they transfer the knowledge of the source domain to the target domain by aligning features, typically by minimizing some form of distributional distance (\eg,~\cite{ganin2017domain,long2018conditional}).
However, in our problem setup, the ImageNet pre-trained model may not be able to learn discriminative features on the source domain due to the limited labels, and this can cause a failure in adaptation to the target domain. In this work, we propose a new self-supervised learning method that exploits unlabeled source and target data to solve the problem by learning not only discriminative but also domain-invariant features. This allows a model to transfer knowledge from the source domain to the target domain even with a few source labels.

Self-supervised learning has recently shown promising results for training deep networks on unlabeled data by defining auxiliary tasks (\eg,~\cite{gidaris2018unsupervised,noroozi2016unsupervised,wu2018unsupervised}). However these approaches do not specifically consider the domain gap issue and cannot ensure domain-invariant features. We therefore propose a new Cross-Domain Self-supervised learning approach (CDS) that utilizes unlabeled data in both the source and target domains for adaptation. We devise two types of self-supervision to extract discriminative and domain-invariant features across both source and target domains: First, we propose \textit{in-domain self-supervision} to learn apparent visual similarity in each domain. This is motivated by recent self-supervised learning~\cite{wu2018unsupervised}, but we apply it in a domain adaptive manner to 
learn discriminative features in each domain. Second, we propose \textit{across-domain self-supervision} to perform cross-domain matching. This objective matches each sample to a neighbor in the other domain while forcing it to be far from unmatched samples. While \textit{in-domain self-supervision} encourages a model to learn discriminative features by separating every instance within a domain, the \textit{across-domain self-supervision} enables better knowledge transfer from the source domain to the target domain by performing instance-to-instance matching across domains. We hypothesize that such features are domain-invariant as well as discriminative, and thus helpful for domain adaptation where there are only a few source labels. Figure~\ref{fig:fig1}-(c) shows that our self-supervised pre-trained model can capture visual similarity as well as semantic similarity across domains compared to an ImageNet pre-trained model (Figure~\ref{fig:fig1}-(b)).




In summary, our work has the following contributions:
\begin{enumerate}[itemsep=1mm]
    \item We propose a new task, unsupervised domain adaptation with few source labels, which is a more practical and challenging task than the conventional unsupervised domain adaption, which assumes many annotated source data.
    \item To address this challenge, we propose a novel Cross-Domain Self-supervised (CDS) 
    method for domain adaptation, which learns discriminative and domain-invariant features without requiring any labels. 
    \item To our knowledge, this is the first method to provide better pre-trained networks against strong baselines including ImageNet pre-trained networks for domain adaptation. We show the effectiveness of our pre-trained networks through extensive experiments. 
\end{enumerate}

	\section{Related Work}

\noindent\textbf{Domain Adaptation.}\quad Traditionally, unsupervised domain adaptation addresses the problem of generalization to a new target domain (no labels) from a fully-labeled source domain. 
Prior domain adaptation methods first extract discriminative features on the source domain guided by source supervision. Then, they align the target features with the source features by:
minimizing maximum mean discrepancy~\cite{long2016unsupervised}, minimizing maximum discrepancy of domain distributions~\cite{saito2018maximum,zhang2019bridging}, feature-level or pixel-level adversarial domain classifier based learning~\cite{ganin2017domain,hoffman2017cycada,long2018conditional,tzeng2014deep}, entropy optimization~\cite{saito2020uni,long2018conditional,saito2019semi},  and finding matching pairs across domains based on optimal transport~\cite{bhushan2018deepjdot,courty2016optimal,shen2018wasserstein,villani2008optimal} or nearest neighbors~\cite{pan2019transferrable,haeusser2017associative}. Some semi-supervised learning techniques such as entropy minimization~\cite{grandvalet2005semi}, pseudo-labeling~\cite{lee2013pseudo}, and Virtual Adversarial Training (VAT)~\cite{miyato2018virtual} have been often used in domain adaptation (\eg,~\cite{lee2019drop,saito2019semi,zou2019confidence}).

The goal of these methods is to completely align feature distributions, such that target features move away from the class decision boundaries learned on the source domain using full supervision. In our problem setup with few source labels, full domain alignment is prone to under-matching due to poor class decision boundaries learned in the source domain. To alleviate the lack of supervision in the source domain, we propose two joint self-supervision losses which induce both domain-invariance and class-discriminative power using only unlabeled data.




In comparison to general semi-supervised learning  (\emph{no target domain})  and semi-supervised domain adaptation~\cite{saito2019semi} (\emph{full source labels and sparse target labels}), our setup (\emph{sparse source labels and no target label}) is far more challenging.

\noindent\textbf{Self-supervised Learning.}\quad Self-supervised learning~\cite{dosovitskiy2015discriminative,gidaris2018unsupervised,noroozi2016unsupervised,wu2018unsupervised} introduces self-supervisory signals for solving pretext tasks. These pretext tasks enable a model to learn generalizable and semantically meaningful features from data for later use in downstream tasks. 
Prior work proposes pretext tasks such as: solving a jigsaw puzzle~\cite{noroozi2016unsupervised}, predicting rotation~\cite{gidaris2018unsupervised}, and Instance Discrimination (ID)~\cite{huang2019unsupervised,wu2018unsupervised}. Instance Discrimination~\cite{wu2018unsupervised} learns an embedding which maps visually similar images closer to each other and far from dissimilar images by classifying an image as its own unique class. Other methods propose to cluster local neighborhoods~\cite{caron2018deep,huang2019unsupervised,saito2020uni,zhuang2019local} within the same domain.

The above methods can provide a pre-trained network for a downstream task, but still assume a large amount of labels in that task for fine-tuning. They also do not consider domain shift between the labeled data and future test data (\eg,~Fig.~\ref{fig:fig1}-(b)). We later show that a network pre-trained on an auxiliary large-scale dataset may not be enough to adapt to our downstream target task due to sparse source labels. On the other hand, our approach aims to learn representations that are generalizable as well as robust to data domain shift without requiring annotation.
 
Some domain adaptation methods~\cite{carlucci2019domain,feng2019self,sun2019unsupervised,xu2019self} directly add existing self-supervised learning objectives (\eg,~\cite{gidaris2018unsupervised,noroozi2016unsupervised}) to improve performance by jointly training with source labels. Saito \etal~\cite{saito2020uni} propose to cluster target features by matching via entropy minimization of feature similarity distribution within the target domain with the help of source supervision. However these methods still rely on full source supervision and the self-supervised learning methods used in their work (\eg,~\cite{gidaris2018unsupervised,noroozi2016unsupervised}) do not promote domain-invariant features on their own (see Sec~\ref{sec:detail_analysis}). In contrast, our method explicitly finds an instance-to-instance matching across domains for domain alignment without any source supervision.





\section{Domain Adaptation with Cross-domain Self-supervised Learning (CDS)}
We explore a new domain adaptation setting where the source domain contains both (sparsely) labeled data $\mathcal{D}_{s}=\left\{\left(\mathbf{x}_{i}^{s}, {y_{i}}^{s}\right)\right\}_{i=1}^{N_{s}}$ as well as unlabeled data $\mathcal{D}_{su}=\left\{\mathbf{x}_{i}^{su}\right\}_{i=1}^{N_{su}}$. We are also provided with unlabeled target domain data $\mathcal{D}_{tu}=\left\{\mathbf{x}_{i}^{tu}\right\}_{i=1}^{N_{tu}}$ that has different data characteristics from the source domain.
Our goal is to train a model using these three data sources, $\mathcal{D}_{s}, \mathcal{D}_{su}$, and $\mathcal{D}_{tu}$, for deploying to the target domain.
This model consists of a CNN feature extractor $F(\cdot)$ followed by a L2 normalization layer~\cite{ranjan2017l2,saito2019semi,wu2018unsupervised}, which outputs a feature vector $\mathbf{f} \in \mathbb{R}^{d}$, and a classifier $C(\cdot)$.
Our domain adaptation framework consists of two stages: (1) Pre-training stage with our cross-domain self-supervised (CDS) learning (Sec.~\ref{sec:pre-training}) and (2) Domain adaptation stage (Sec.~\ref{sec:da}). 
The goal of the first stage is to obtain the pre-trained weights that are robust to domain-shift and efficiently generalizable for later domain adaptation.
Up to this step, no label is required; \ie, a purely self-supervised step. In the second stage, with our pretrained weight, we apply existing domain adaptation methods such as~\cite{ganin2017domain,long2018conditional,saito2019semi} with few source labels.


We aim to learn class-discriminative and domain-invariant features in different domains with CDS. The previous self-supervised learning method~\cite{wu2018unsupervised} learns visual similarity where a model embeds visually \textit{similar-looking} images nearby while being far away from \textit{dissimilar-looking} images. However, in domain adaptation, the same class images may look very different due to domain gap, so that visual similarity learning alone does not ensure semantic similarity and domain-invariance between domains.

\begin{figure}[t]
	\centering
	\includegraphics[width=\linewidth]{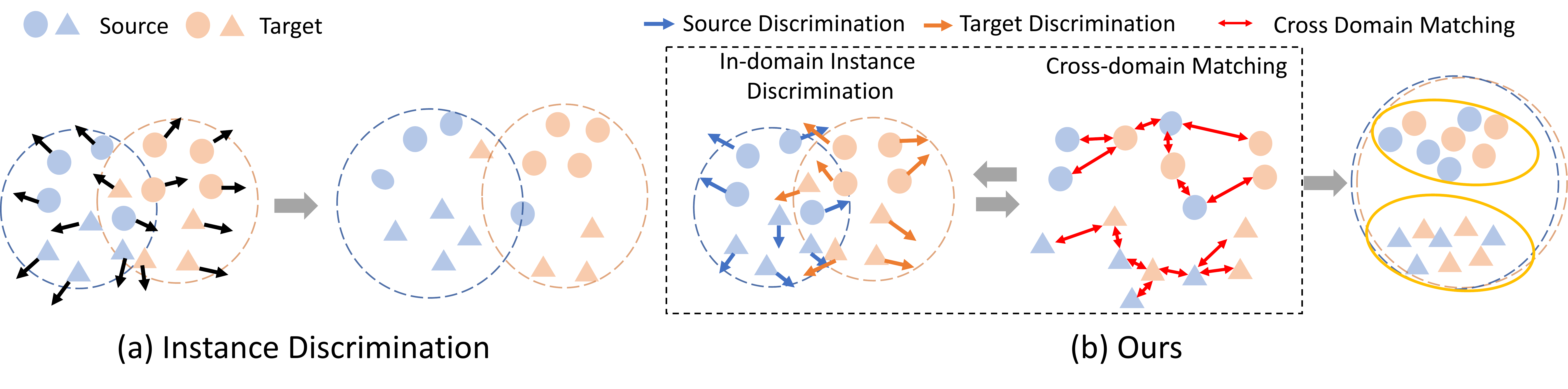}
	\vspace{-2.5em}
	\caption{\small
		 Comparison of instance discrimination~\cite{wu2018unsupervised} and our cross-domain self-supervised learning: (a) Instance discrimination distinguishes every feature from all the others without considering the domain gap, so that features of different domains are unlikely to be embedded close together. (b) In order to reduce the domain gap, our method jointly uses in-domain instance discrimination and cross-domain alignment to learn features that are domain-invariant as well as discriminative (best viewed in color). 
		}
	\label{fig:selfsup}
\end{figure}

Our CDS consists of two objectives: (1) learning visual similarity with in-domain supervision and (2) cross-domain matching with across-domain supervision. Figure~\ref{fig:selfsup} illustrates the differences between the existing and our method.


\smallskip
\subsection{Pre-training with Cross-domain Self-supervision (CDS)}
\label{sec:pre-training}

\begin{figure}[t]
	\centering
	\includegraphics[width=1.0\linewidth]{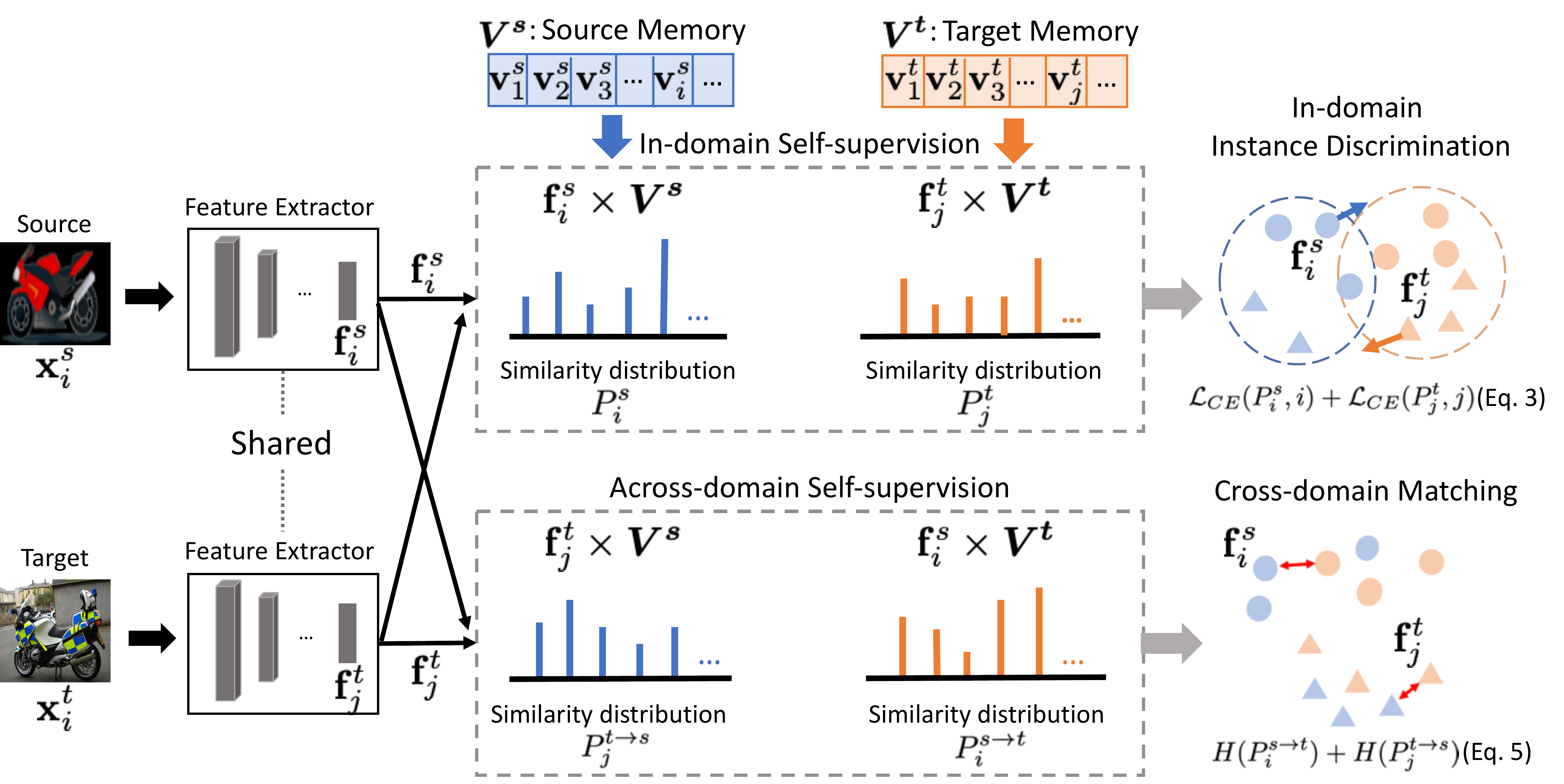}
	\vspace{-1.0em}
	\caption{
		An overview of the pre-training stage with our cross-domain self-supervised learning, which extracts discriminative and domain-invariant features on the source and target domain. In the \textit{in-domain self-supervision}, we measure the similarity of features in-domain, and then perform in-domain instance discrimination to learn visual similarity in each domain. In the \textit{across-domain self-supervision}, we measure similarity between a feature and cross-domain features from the cross-domain memory bank and then minimize the entropy for cross-domain matching (best viewed in color).}
	\label{fig:selfsup2}
\end{figure}

\noindent\textbf{In-domain Self-supervision.}\quad
The goal of this part is to learn a discriminative feature extractor which captures apparent visual similarity for two different domains with in-domain supervision. 
For a single domain-only, Instance Discrimination~\cite{wu2018unsupervised} (ID) is proposed to learn visual similarity by imposing a distinctive unique class to every image instance and by training a model such that the images are classified to its own instance identity by \textit{treating all the other images as negative pairs}. By ID, they hypothesize that a model can discover the underlying class-to-class semantic similarity (\ie,~class-discriminative) which are helpful for a recognition task as shown in~\cite{he2019momentum,wu2018unsupervised}.

A naive deployment of ID to the domain adaptation scenario, treating all the other samples as negatives against a given query sample without distinguishing domains, has several drawbacks. 
ID can encourage two images of the same class but in different domains to embed far from each other due to the different visual characteristics of their respective domains (see Sec. \ref{sec:detail_analysis} for analysis). 
Analogously, ID can incorrectly embed images of different categories belonging to different domains nearby each other due to spuriously shared visual characteristics as in Fig.~\ref{fig:fig1}-(b). Also, ID discards clustering effects by making every instance far away as shown in~\cite{huang2019unsupervised}, which does not learn features both class-discriminative and domain-invariant; thus, unfavorable  for domain adaptation.

In order to alleviate these problems of ID for domain adaptation, 
we propose to use in-domain instance discrimination, where negative pairs are from in-domain samples in each domain-specific memory bank defined below. This aims to prevent learning wrong visual clues from the other domain (\eg,~Fig.~\ref{fig:fig1}-(a)) and discriminating two domain features, as illustrated in Fig.~\ref{fig:selfsup}-(a).

We first initialize the source and target memory banks $\bm{V^s}$ and  $\bm{V^t}$ with all the source and target features, respectively, from the feature extractor $F(\cdot)$,
\begin{equation}
\bm{V^s} = [\mathbf{v}_1^s, \cdots, \mathbf{v}_{(N_s+N_{su})}^s], \quad \bm{V^t} = [\mathbf{v}_1^t, \mathbf{v}_2^t, \cdots, \mathbf{v}_{N_{tu}}^t],
\end{equation}
where $\mathbf{v}_i$ is the feature vector of the image $\mathbf{x}_i$ , \ie,~$\mathbf{v}_i^s = F(\mathbf{x}_i^s)$. After this initialization, the memory bank features will be updated with a momentum in every batch (described in later section).

Using the feature extractor $F(\cdot)$, we obtain feature vectors $\mathbf{f}^s=F(\mathbf{x}_i^s)$ and $\mathbf{f}^t=F(\mathbf{x}_j^t)$ from a source image $\mathbf{x}_i^s \in B^s$ and a target image $\mathbf{x}_j^t \in B^t$ in respective batches, $B^s \cup B^t$. To perform in-domain instance discrimination, we compute the similarity distributions $P^{s}_i $ and $P^{ t}_j$ by measuring the pairwise similarities between features and the corresponding memory bank (the top row of Fig.~\ref{fig:selfsup2}),
\begin{equation}
P^{s}_i = \frac{\exp ((\mathbf{v}_i^s)^\top \mathbf{f}_i^s / \tau)}{\sum_{k=1}^{N_s+ N_{su}} \exp(({\mathbf{v}_k^s})^\top \mathbf{f}_i^s / \tau)},  \quad
P^{ t}_j = \frac{\exp ((\mathbf{v}_j^t)^\top \mathbf{f}_j^t / \tau)}{\sum_{k=1}^{N_{tu}} \exp((\mathbf{v}_k^t)^\top \mathbf{f}_j^t) / \tau)},
\end{equation}
where the temperature parameter $\tau$ determines the concentration level of the similarity distribution~\cite{hinton2015distilling}. Finally, we perform the in-domain instance discrimination by minimizing the averaged cross entropy losses of each domain in a batch:
\begin{equation}
\textstyle
\mathcal{L}_{W-INS} = \frac{1}{|B^s+B^t|}(\sum\nolimits_{i\in B^s} \mathcal{L}_{CE}(P_i^s, i) + \sum\nolimits_{j \in B^t}\mathcal{L}_{CE}(P_j^t, j)),
\label{eq:d_ins}
\end{equation}
where $i$ and $j$ denote the unique index of the samples of $x_i$ and $x_j$. 

\smallskip

\noindent\textbf{Across-domain Self-supervision.}\quad
To explicitly ensure domain aligned and discriminative features between the two different domains, we perform cross-domain feature matching.

Prior work accomplished this by using an adversarial domain classifier~\cite{ganin2017domain} or Mean Maximum Discrepancy~\cite{long2016unsupervised} to align two domain feature distributions. Optimal transport~\cite{villani2008optimal} is often used to find a matching pair of two distributions, but this scales poorly~\cite{courty2016optimal} and is limited to find a matching in a batch~\cite{bhushan2018deepjdot}, while we find a matching globally in all cross-domain samples by using ``cached features'' in the memory bank instaed of ``live'' features to compute the maching. In addition, these methods focus on minimizing the gap between distributions of two domains but do not consider 
class-class semantic similarity of the source and target domains and lose class-discriminative power. Instead, our method globally discovers negative matchings as well as a positive matching to ensure class-discriminative features in different domains. This can be achieved by matching a sample in one domain with samples in another domain while enforcing the sample far from unmatched samples. To find a match, we minimize the entropy of the pairwise similarity distribution between a feature in one domain and features in the other domain memory bank. 

Given the feature vectors (queries), $\mathbf{f}^s{=}F(\mathbf{x}_i^s)$ and $\mathbf{f}^t{=}F(\mathbf{x}_j^t)$, on the respective source and target images $\mathbf{x}_i^s \in B^s$ and $\mathbf{x}_j^t \in B^t$ in a batch ($B^s \cup B^t$),
we first measure across-domain pairwise similarities between the feature and the across-domain memory bank features  
(the bottom row of Fig.~\ref{fig:selfsup2}) as
\begin{equation}
P^{s \veryshortarrow t}_{i', i} = \frac{\exp ( (\mathbf{v}_{i'}^t)^\top  \mathbf{f}_i^s / \tau)}{\sum_{k=1}^{N_{tu}} \exp(({\mathbf{v}_k^t})^\top \mathbf{f}_i^s / \tau)},  \quad
P^{ t \veryshortarrow s}_{j',j} = \frac{\exp ((\mathbf{v}_{j'}^s)^\top \mathbf{f}_j^t / \tau)}{\sum_{k=1}^{N_s+ N_{su}} \exp((\mathbf{v}_k^s)^\top \mathbf{f}_j^t / \tau)}\textrm{.}
\end{equation}

Then we minimize the averaged entropy of the similarity distributions in a batch, which clusters source and target features and encourages distribution alignment, as follows:

\begin{equation}
\textstyle
\mathcal{L}_{CDM} = \frac{1}{|B_s + B_t|} (\sum\nolimits_{i \in B^s} H(P^{s \veryshortarrow t}_i) + \sum\nolimits_{j \in B^t} H(P^{t \veryshortarrow s}_j)) \textrm{, where}
\label{eq:cda}
\end{equation}
\begin{equation}
 H(P^{s \veryshortarrow t}_i) =  - {\displaystyle \sum_{i'}^{{\scriptstyle N_{tu}}}} P^{s \veryshortarrow t}_{i',i}
 \log {P^{s \veryshortarrow t}_{i',i}},{\scriptstyle \quad}H(P^{t \veryshortarrow s}_j) = - {\displaystyle \sum_{j'}^{\scriptstyle {N_s+N_{su}}}} P^{t \veryshortarrow s}_{j',j} \log P^{t \veryshortarrow s}_{j',j}
 \textrm{.}\nonumber
\end{equation}

\noindent Since entropy minimization encourages a model to make a confident prediction, the model chooses a sample to match and enforces the query feature (\ie, $\mathbf{f}_i^s$ or $\mathbf{f}_i^t$ ) to be closer to the matched sample. At the same time, The model enforces the query feature to be far from all the other unmatched examples in another domain, which learns class-discriminative features across domains.

  The overall objective for CDS is to minimize: 
\begin{equation}
\mathcal{L}_{CDS} =\mathcal{L}_{W-INS} +\mathcal{L}_{CDM}.
\end{equation}
After updating the model with the losses in Eqs.~(\ref{eq:d_ins}) and~(\ref{eq:cda}), we update the memory banks with the features in the batch with a momentum $\eta$ following~\cite{wu2018unsupervised}:
\begin{equation}
\forall i \in B^s,
\mathbf{v}_i^s =(1-\eta) 
\mathbf{v}_i^s+\eta \mathbf{f}_i^s,
\quad
\forall j \in B^t ,
\mathbf{v}_j^t =(1-\eta) 
\mathbf{v}_j^t+\eta 
\mathbf{f}_j^t,
\label{eq:memory_update}
\end{equation}

\subsection{Domain Adaptation}
\label{sec:da}
 This stage transfers knowledge of the source domain to the unlabeled target domain with few source labels. After the pre-training stage (Sec. \ref{sec:pre-training}), we have a network pretrained with
 our cross-domain self-supervised learning. Then we apply learning objective functions of a domain adaptation method ($\mathcal{L}_{DA}$) with labeled source and unlabeled target examples. For the unlabeled source domain, we apply a semi-supervised learning method ($\mathcal{L}_{su}$) (\eg,~entropy minimization~\cite{grandvalet2005semi}). Finally, we optimize the loss function $\mathcal{L}$:
\begin{equation}
\mathcal{L} = \mathcal{L}_{DA}(\mathcal{D}_{s}, \mathcal{D}_{tu}) + \lambda \mathcal{L}_{su}(\mathcal{D}_{su})
\end{equation}
where $\lambda$ is the hyper-parameter that controls the importance of the semi-supervised learning loss. 
Later, we will show the effectiveness of our pre-trained network by exploring
various types of domain adaptation methods and semi-supervised learning methods in our experiments.


	\section{Experiments}
We evaluate our method (CDS) on our new domain adaptation setting with few source labels. We treat some portion of source data as labeled and others as unlabeled. We explain our evaluation setup in Sec~\ref{sec:dataset}, and our implementation details in Sec.~\ref{sec:implementation}.
We report results on the new domain adaptation setting in Sec.~\ref{sec:domain_few_labels}. We analyze the impact of semi-supervised learning methods as well as the comparison with other self-supervised learning baselines in our domain adaptation setting in Sec.~\ref{sec:ablation_semi}. 
We additionally assess the quality of our representation in several aspects in Sec.~\ref{sec:detail_analysis}. In Sec.~\ref{sec:traditional_domain_adaptation}, we show that CDS is also effective on the traditional domain adaptation settings. Additional results can be found in the supplementary material.

\subsection{Experiment Setting}
\label{sec:dataset}\label{sec:implementation}
\noindent\textbf{Dataset.}\quad
Since we propose a new task, there is no benchmark that is specifically designed for our task.
We utilize three standard domain adaptation benchmarks for evaluation in our task. 
Table~\ref{exp:dataset} shows the overall statistics of the datasets and the number of labeled source examples used in our experiments. We holdout a majority of source labels during training to mimic the sparse source label regime in our proposed task.
We ensure that each class has at least one labeled example. Since Office is a relatively easier dataset compared to Office-Home and VisDA, we experiment 1-shot and 3-shots source labels per class. 

\begin{table}[t]
\vspace{-3mm}
\resizebox{\textwidth}{!}{\begin{tabular}{l|c|c|c|c|c|c|c|c|c}
\hline
Dataset & \multicolumn{3}{c|}{Office~\cite{saenko2010adapting}} & \multicolumn{4}{c|}{Office-Home~\cite{venkateswara2017deep}} & \multicolumn{2}{c}{VisDA~\cite{peng2017visda}} \\
\hline
Domain & Amazon (A) & Dslr (D) & Webcam (W) & Art (Ar) & Clipart (Cl) & Product (Pr) & Real (Rw) & Synthetic (Syn) & Real (Rw) \\
\cline{2-10}
\# total images & 2817 & 498 & 795 & 2427 & 4365 & 4439 & 4357 & 152K & 55K \\
\cline{2-10}
\# labeled images & \multicolumn{3}{c|}{1-shot and 3-shots labeled source} & \multicolumn{4}{c|}{3\%, 6\%, and 12\%
labeled source} & \multicolumn{2}{c}{1\% and 0.1\% labeled source} \\
\cline{2-10}
\# classes & \multicolumn{3}{c|}{31} & \multicolumn{4}{c|}{65} & \multicolumn{2}{c}{12}  \\
\hline
\end{tabular}}
\vspace{0.05em}
\caption{Dataset statistics used in our experiments.}
\label{exp:dataset}
\vspace{-2.0em}
\end{table}

\begin{table}[t]	
	\centering
	\resizebox{\textwidth}{!}{	\begin{tabular}{c||c|c|c|c|c|c|c|c}
		\toprule[1.0pt]

			\multirow{2}{*}{Adapt}&\multirow{2}{*}{Pretrain} & \multicolumn{6}{c}{Office: Target Acc. (\%) on 1-shot / 3-shots } \\
			\cline{3-9}
           & & A$\rightarrow$D & A$\rightarrow$W & D$\rightarrow$ A & D$\rightarrow$W & W$\rightarrow$A & W$\rightarrow$D & AVG\\
			\hline
			SO& IN~\cite{russakovsky2015imagenet}  & 30.5 / 50.6 & 25.9 / 55.5 & 35.9 / 51.9 & 67.9 / 83.9 & 36.4 / 50.7 & 49.4 / 85.9 & 41.0 / 63. 1\\
			DANN &IN & 32.5 / 57.6 & 37.2 / 54.1 & 36.9 / 54.1 & 70.1 / 87.4 & 43.0 / 51.4 & 58.8 / 89.4 & 46.4 / 65.7  \\
			\hline
			\multirow{2}{*}{CDAN} &IN & 31.5 / 68.3 & 26.4 / 71.8 & 39.1 / 57.3 & 70.4 / 88.2 & 37.5 / 61.5 & 61.9 / 93.8 &  44.5 / 73.5\\
			&CDS & \textbf{53.8} /\textbf{ 78.1} & \textbf{65.6} / \textbf{79.8} & \textbf{59.5} / \textbf{70.7} & \textbf{83.0} / \textbf{93.2} & \textbf{57.4} / \textbf{64.5} & \textbf{77.1} / \textbf{97.4} & \textbf{66.1} / \textbf{80.6}\\
			\hline
				\multirow{2}{*}{MME}  &IN  & 37.6 / 69.5 & 42.5 / 68.3 & 48.6 / 66.7 & 73.5 / 89.8 & 47.2 / 63.2 & 62.4 / 95.4 & 52.0 / 74.1 \\
		&CDS & \textbf{54.4} / \textbf{75.7} & \textbf{57.2} / \textbf{77.2 }&\textbf{ 62.8 }/ \textbf{69.7 }& \textbf{84.9} / \textbf{92.1 }& \textbf{62.6} / \textbf{69.9}& \textbf{77.7}/ \textbf{95.4} & \textbf{65.3} / \textbf{80.0}\\
					  \bottomrule[1.0pt]
	\end{tabular}}
	\vspace{0.1em}
	\caption{Target  accuracy (\%)  on 1-shot and 3-shots per class on the Office dataset. Entropy minimization is applied to unlabeled source examples. The first column (Adapt) refers to the domain adaptation methods and the second column (Pretrain) refers to pre-training methods used in these experiments. IN denotes ImageNet-pretrained and CDS denotes our cross-domain self-supervised learning.}
	\label{exp:semi_office}
	\vspace{-1.5em}
\end{table}


\vspace{1mm}
\noindent\textbf{Implementation Details.}\quad
Our CDS is implemented in PyTorch~\cite{paszke2017automatic}. We use a ResNet-50~\cite{he2016deep} pre-trained on ImageNet followed by a randomly initialized linear layer and a $L_2$ normalization layer as an initial feature extractor for all experiments. 
\textit{In the pre-training stage} with CDS, we use SGD with the moment parameter $0.9$, a learning rate of 0.003, a batch size of $64$, weight decay rate $5e^{-4}$. As for the parameters $\tau$ and $\eta$, we set $\tau=0.05$ and  $\eta=0.5$.

\subsection{Domain Adaptation with Few Source Labels}
\label{sec:domain_few_labels}

\noindent\textbf{Setup.}\quad 
We compare our pre-trained weights (CDS) with an ImageNet pre-trained weights in the proposed domain adaptation task. To show that CDS is widely applicable to domain adaptation methods, we consider the variety of seminal domain adaptation methods (\ie, $\mathcal{L}_{DA}$ in Eq.~(\ref{eq:cda})):
\textbf{SO (Source-only)}, \textbf{DANN}~\cite{ganin2017domain}, \textbf{CDAN}~\cite{long2018conditional}, \textbf{MME}~\cite{saito2019semi}. DANN and CDAN are based on a domain classifier and regarded as standard
baselines. We also compare with
other types of domain adaptation baselines, which are based on the maximum discrepancy based methods MCD~\cite{saito2018maximum} and MDD~\cite{zhang2019bridging} for additional analysis. We use the original authors' codes~\cite{long2018conditional,saito2019semi,saito2018maximum,zhang2019bridging}. We follow the experiment setup suggested by Saito~\etal~\cite{saito2019semi}, where we use the same validation set (3-shots per class) to select the best accuracy on the target domain and use early stopping.  For all methods, entropy minimization (Ent) is applied for the unlabeled source examples.
We choose the hyper-parameter $\lambda \in \{0.01, 0.05, 0.1, 0.2 ,0.3\}$  for Ent ($\mathcal{L}_{su}$ in Eq.~(\ref{eq:cda})). 




\begin{table}[t]
	\resizebox{\textwidth}{!}{\begin{tabular}{c||c|c|c|c|c|c|c|c|c|c|c|c|c|c}
	\toprule[1.0pt]
			\multirow{2}{*}{Adapt}	&	\multirow{2}{*}{Pretrain} & \multicolumn{13}{c}{Office-Home: Target Acc. (\%)} \\
			\cline{3-15}
		 & & Ar$\veryshortarrow$Cl &Ar$\veryshortarrow$Pr&Ar$\veryshortarrow$Rw&Cl$\veryshortarrow$Ar&Cl$\veryshortarrow$Pr&Cl$\veryshortarrow$Rw&Pr$\veryshortarrow$Ar&Pr$\veryshortarrow$Cl&Pr$\veryshortarrow$Rw&Rw$\veryshortarrow$Ar&Rw $\veryshortarrow$Cl&Rw$\veryshortarrow$Pr& AVG \\
			\hline
		
			\multicolumn{15}{c}{\textbf{(a) 12\% labeled source}}    \\	\hline
			SO & IN & 25.9 & 42.1 & 52.1 & 32.3 & 36.0 & 38.9 & 37.5 & 29.1 & 56.0 & 50.5 & 34.8 & 63.7 & 41.6 \\
			DANN & IN & 27.6 & 42.3 & 49.4 & 27.9 & 39.6 & 39.9 & 37.7 & 28.8 & 58.7 & 46.7 & 36.2 & 63.6 & 41.6 \\
			\hline
			\multirow{2}{*}{CDAN} & IN & 28.1 & 42.4 & 52.8 & 39.3 & 50.9 & 49.6 & 48.1 & 39.0 & 67.7 & 58.1 & 41.7 & 73.3 & 49.2 \\
			 &CDS & \textbf{42.2} & \textbf{55.9} & \textbf{64.6} & \textbf{54.1} & \textbf{60.6} & \textbf{67.0} & \textbf{54.8} & \textbf{48.4} & \textbf{74.0} & \textbf{67.0} & \textbf{48.9} & \textbf{75.9} & \textbf{59.5} \\
			\hline
			\multirow{2}{*}{MME}& IN & 41.7 & 49.8 & 59.4 & 50.3 & 52.6 & 54.9 & 54.6 & \textbf{51.3} & 71.7 & 65.7 & 51.4 & 76.1 & 56.6 \\
			 &CDS & \textbf{45.6} & \textbf{57.3} & \textbf{67.0} & \textbf{57.4} & \textbf{58.5} & \textbf{64.6} & \textbf{58.7} & 51.2 & \textbf{75.2} & \textbf{68.6} & \textbf{57.4} & \textbf{76.3} & \textbf{61.5} \\
			\hline
			\hline
			\multicolumn{15}{c}{\textbf{(b) 6\% labeled source}}  \\	\hline
			SO & IN & 22.8 & 36.2 & 45.2 & 26.3 & 30.4 & 33.9 & 33.3 & 28.2 & 52.7 & 45.0 & 30.5 & 58.3 & 36.9 \\
			DANN & IN & 22.4 & 32.9 & 43.5 & 23.2 & 30.9 & 33.3 & 33.2 & 26.9 & 54.6 & 46.8 & 32.7 & 55.1 & 36.3 \\
			\hline
			\multirow{2}{*}{CDAN}& IN & 23.1 & 35.5 & 49.2 & 26.1 & 39.2 & 43.8 & 44.7 & 33.8 & 61.7 & 55.1 & 34.7 & 67.9 & 42.9 \\
			 &CDS& \textbf{39.0} & \textbf{51.3} & \textbf{63.1} & \textbf{51.0} & \textbf{55.0} & \textbf{63.6} & \textbf{57.8} & \textbf{45.9} & \textbf{72.8} & \textbf{65.8} & \textbf{50.4} & \textbf{73.5} & \textbf{57.4} \\
			\hline
			\multirow{2}{*}{MME} &IN & 37.2 & 42.4 & 50.9 & 46.1 & 46.6 & 49.1 & 53.5 & 45.6 & 67.2 & 63.4 & 48.1 & 71.2 & 51.8 \\
			 &CDS & \textbf{44.1} & \textbf{51.6} & \textbf{63.3} & \textbf{53.9} & \textbf{55.2} & \textbf{62.0} & \textbf{56.5} & \textbf{46.6} & \textbf{70.9} & \textbf{67.7} & \textbf{54.7} & \textbf{74.7} & \textbf{58.4} \\
			\hline
			\hline
			\multicolumn{15}{c}{\textbf{(c) 3\% labeled source}}  \\	\hline
			SO & IN & 20.9 & 31.5 & 38.6 & 19.7 & 24.8 & 22.6 & 32.9 & 25.3 & 48.1 & 40.8 & 24.1 & 48.6 & 31.5 \\
			DANN &IN & 19.5 & 30.2 & 38.1 & 18.1 & 21.8 & 24.2 & 31.6 & 23.5 & 48.1 & 40.7 & 28.1 & 50.2 & 31.2 \\
			\hline
			\multirow{2}{*}{CDAN} &IN & 20.6 & 31.4 & 41.2 & 20.6 & 24.9 & 30.6 & 33.5 & 26.5 & 56.7 & 46.9 & 29.5 & 48.4 & 34.2 \\
			 &CDS  & \textbf{37.7} & \textbf{49.2} & \textbf{56.5} & \textbf{49.8} & \textbf{51.9} & \textbf{55.9} & \textbf{50.0} & \textbf{42.3} & \textbf{68.1} & \textbf{63.1} & \textbf{48.7} & \textbf{67.5} & \textbf{53.4} \\
			\hline
			\multirow{2}{*}{MME}  &IN & 31.2 & 35.2 & 40.2 & 37.3 & 39.5 & 37.4 & 48.7 & 42.9 & 60.9 & 59.3 & 46.4 & 58.6 & 44.8 \\
			 & CDS & \textbf{41.7} & \textbf{49.4} & \textbf{57.8} & \textbf{51.8} & \textbf{52.3} & \textbf{55.9} & \textbf{54.3} & \textbf{46.2} & \textbf{69.0} & \textbf{65.6} & \textbf{52.2} & \textbf{68.2} & \textbf{55.4} \\
			  \bottomrule[1.0pt]

	\end{tabular}}
\vspace{0.1em}
	\caption{Target  accuracy (\%)  on the Office-Home dataset under the different number of labeled source examples. Entropy minimization is applied to unlabeled source examples.}
	\label{exp:semi_office_home}
	\vspace{-1.0em}
\end{table}

\vspace{2mm}\noindent\textbf{Adaptation Results with Few Source Labels.}\quad 
Tables~\ref{exp:semi_office},~\ref{exp:semi_office_home}, and~\ref{exp:semi_visda} show the comparison of our pre-trained weights (denoted by CDS) with ImageNet pre-trained weights (denoted by IN) on Office, Office-Home, and VisDA. CDS improves the performance in all cases except the setting Pr$\veryshortarrow$Cl on Office-Home, with $12\%$ labels where CDS shows a comparable accuracy. As the number of labeled examples decreases, CDS shows higher performance gains against the baselines. In Table~\ref{exp:semi_visda} on VisDA, the performance gain tends to be larger with the weak domain adaptation methods. These results show that our self-supervised learning scheme is more effective than just naive adaptation of ImageNet pre-trained weights, which is generally used in domain adaptation works. 


\begin{SCtable}[][t]
\centering
\resizebox{0.45\textwidth}{!}{\begin{tabular}{lr@{\hspace{3mm}}|@{\hspace{3mm}}lr@{\hspace{3mm}}|@{\hspace{3mm}}lr@{\hspace{3mm}}|@{\hspace{3mm}}lr}
	\toprule[1.0pt]

\multicolumn{8}{c}{VisDA: Target Acc. (\%)}\\
\hline
 \multicolumn{2}{c|@{\hspace{3mm}}}{SO} & \multicolumn{2}{c|@{\hspace{3mm}}}{DANN} & \multicolumn{2}{c|@{\hspace{3mm}}}{CDAN}  & \multicolumn{2}{c}{MME}  \\\hline
  \multicolumn{8}{c}{(a) 1\% labeled source} \\ \hline 
 IN & 38.2 & IN & 50.2  & IN & 58.1 & IN & 66.1 \\
 CDS & \textbf{47.4} & CDS & \textbf{63.7} & CDS& \textbf{69.1} & Ours & \textbf{69.4} \\\hline \hline
  \multicolumn{8}{c}{(b) 0.1\% labeled source} \\\hline
 IN & 37.1 & IN & 44.49 & IN & 57.7 & IN & 54.0 \\
 CDS & \textbf{45.5} & CDS & \textbf{62.9} & CDS & \textbf{69.0} & Ours & \textbf{62.5} \\
\bottomrule[1.0pt]

\end{tabular}}
\vspace{0.1em}

\caption{Target accuracy (AVG \%) from Synthetic to Real setting. A model in each method is initialized with ImageNet-pretrained (IN) or our self-supervised learning (CDS). 
}
\label{exp:semi_visda}
\vspace{-2.0em}
\end{SCtable}

For the maximum classifier discrepancy based methods (MCD and MDD), we observe that these methods tend to have decreasing target accuracy and to collapse to random predictions when there are only few source labels (\eg,~3.2\%  Acc in 1-shot web-to-amazon setting). 
We postulate that the classifiers failed to learn sharp class decision boundaries due to the lack of labeled source examples, which seems insufficient to obtain generalizable inductive bias for maximum discrepancy classifiers. Detailed results can be found in the supplementary material.

\begin{table}[t]
 	\resizebox{\textwidth}{!}{	
	\begin{tabular}{rlc|c|c|c|c|c|c|c}
		\toprule[1.0pt]
			& & & \multicolumn{6}{c}{Office: Target Acc. (\%) / Source Acc. (\%) on 1-shot} \\
			\cline{4-10}
			& Adapt & Semi-sup. & A$\rightarrow$D & A$\rightarrow$W & D$\rightarrow$ A & D$\rightarrow$W & W$\rightarrow$A & W$\rightarrow$D & AVG\\
			\hline
			\hline
			&  \multicolumn{9}{c}{ (a) ImageNet pre-trained}\\
			\hline
			&SO & - & 28.3 / 39.8 & 31.6 / 42.0 & 34.8 / 71.3 & 64.5 / 68.5 & 37.0 / 64.0 & 56.8 / 64.1 & 42.2 / 58.3\\[0.5mm]
			&SO & ENT+VAT & 27.9 / 38.4 & 31.5 / 39.3 & 37.7 / 69.6 & 69.9 / 71.1 & 36.5 / 62.7 & 48.6 / 59.2 & 42.0 / 56.7\\
			\hline
			&	MME & -& 50.4 / 36.3 & 49.6 / 34.5 & 47.2 / 67.9 & 78.2 / 77.9 & 46.1 / 66.9 & 66.9 / 67.9 & 56.4 /  58.7\\[0.5mm]
			&MME&ENT+VAT & 40.1 / 47.6 & 41.6 / 42.2 & 48.1 / 77.7 & 63.7 / 76.7 & 46.8 / 70.4 & 62.2 / 67.9 & 50.4 / 63.8\\
			\hline
			\hline
			&  \multicolumn{9}{c}{ (b) \textbf{CDS}}\\\hline
			&SO &- & 52.2 / 56.9 & 54.6 / 54.3 & 51.3 / 78.6 & 78.5 / 79.9 & 55.3 / 73.3 & 71.5 / 72.9 & 60.6 /  69.3 \\[0.5mm]
			&SO&ENT+VAT & 53.4 / 57.9 & 56.7 / 56.9 & 56.2 / 80.7 & 80.8 / 80.7 & 56.2 / 76.6 & 73.5 / 74.9 & 62.8 / 71.3 \\
			\hline
			&MME &- & 51.2 / 35.1 & 56.9 / 39.3 & 58.0 / 82.2 & 80.3 / 80.3 & 58.6 / 76.7 & 70.3 / 72.5 & 62.6 / 64.4 \\[0.5mm]
			&MME&ENT+VAT  & 53.2 / 58.5 & 58.1 / 54.1 & 69.5 / 88.1 & 82.5 / 85.0 & 62.2 / 76.6 & 76.9 / 80.9 & 67.1 / 73.9\\
			\bottomrule[1.0pt]
	\end{tabular}}
    \vspace{0.1em}
	\caption{Target and source accuracy (\%)  on unlabeled samples with a semi-supervised learning method. Different from the ImageNet pre-trained network, CDS provides domain-invariant features, so that additional semi-supervised learning objectives on the source domain can help to improve target accuracy.}
	\label{tab:office_1_shot_semi}
	\vspace{-2.0em}
\end{table}

\vspace{2mm}\noindent\textbf{Effects of Different Semi-supervised Learning Methods.} \quad
\label{sec:ablation_semi}
In order to further analyze the effect of semi-supervised learning in our proposed task, we report comparisons with the following semi-supervised learning method:
(1) Labeled source only, (2) Entropy Minimization + Virtual Adversarial Training (Ent+VAT)~\cite{miyato2018virtual,oliver2018realistic,zhai2019s4l}. VAT first generates adversarial perturbations which change the output of the model significantly and train a model to be robust to the perturbations. These two methods are also widely used in the domain adaptation~\cite{long2018conditional,saito2019semi,lee2019drop}. We choose the hyper-parameter $\lambda \in \{0.01, 0.05, 0.1, 0.2 ,0.3\}$  for Ent and $\lambda_{eps} \in\{5, 10,15, 20, 25, 30\}$ for VAT on the dslr-to-amazon setting. More results can be found in supplementary. 

Table~\ref{tab:office_1_shot_semi} shows the effect of the semi-supervised learning methods. We measure accuracy on the unlabeled target and unlabeled source domain. In comparison to traditional semi-supervised learning, (\eg,~\cite{miyato2018virtual,grandvalet2005semi}), we use ImageNet pre-trained weights by following domain adaptation works and the number of labeled and unlabeled examples is very low (\eg,~The number of data in the dslr domain is only 498).  In Table~\ref{tab:office_1_shot_semi}-(a), the semi-supervised learning does not help much on the source accuracy for Source-only (SO). This similar behavior is observed in~\cite{saito2019semi}, entropy minimization can harm accuracy when there are only few labels. For MME, the semi-supervised learning shows significant improvements on the source accuracy under the various settings. However, it is interesting to see that the increased source accuracy harms the target accuracy on the A$\rightarrow$D and A$\rightarrow$W settings in Table~\ref{tab:office_1_shot_semi}-(a), which suggesting that source accuracy is not necessarily consistent with target accuracy. In Table~\ref{tab:office_1_shot_semi}-(b), CDS can improve the source and target accuracy at the same time on the A$\rightarrow$D and A$\rightarrow$W settings with domain-invariant features. Lastly, we observe that, with CDS, the performance gap of the target accuracy between SO and MME is significantly and consistently smaller than
that of ImageNet pre-trained network $14.2\% \rightarrow 4.3\%$ in AVG). These results show that CDS can produce features favorable for domain adaptation.

\vspace{2mm}\noindent
\textbf{Self-supervised Learning Method Comparison.\quad}
For further analysis, we compare with other self-supervised learning baselines: Instance Discrimination (ID), RotNet~\cite{gidaris2018unsupervised} and Jigsaw Puzzle~\cite{noroozi2016unsupervised}. 
We integrate domain alignment with an adversarial domain classifier to build a fair yet commonly used baseline. Table~\ref{exp:comparison_selfsup} shows the comparison of ours with the commonly used self-supervised learning baselines. When applying self-supervised learning, we apply it to the union set of the source and target domain samples.

\begin{table}[t]
	\centering
	\setlength{\tabcolsep}{7pt}
	\resizebox{0.9\textwidth}{!}{\begin{tabular}{l|c|c|c|c|c|c|c|c}
				\toprule[1.0pt]
			\multirow{3}{*}{Pretrain} &  \multicolumn{4}{c|}{ Office: D$\veryshortarrow$A } &   
			\multicolumn{4}{c}{Office-Home: Rw$\veryshortarrow$Cl} \\
			\cline{2-9}
			& \multicolumn{2}{c|}{Domain Adaptation } &    \multicolumn{2}{c|}{Feature Analysis} &  \multicolumn{2}{c|}{Domain Adaptation } &    \multicolumn{2}{c}{Feature Analysis } \\
			\cline{2-9}
			 & CDAN  &MME & Linear & kNN &  CDAN  &MME & Linear & kNN      \\
			\hline
			ImageNet only & 39.1 & 48.6 & 62.7 & 52.9   &29.5 & 46.4 & 37.6 & 37.1    \\
			ID  & 43.5 & 46.9 & 63.8 &  60.8 & 28.4 & 46.8   & 42.9  & 44.7  \\
			DC & 40.7 & 51.0  & 64.2 & 45.7  & 29.8 & 47.1  & 39.0 &  33.1  \\
			ID+DC  & 40.6 & 48.5 &  56.8  &62.7  & 28.0 & 43.8  & 43.8 & 44.8 \\
			Jigsaw~\cite{noroozi2016unsupervised}   &41.1 & 50.7 &52.2 & 26.8   & 29.3 &  43.9  & 38.8  & 24.4\\
			Rotation~\cite{gidaris2018unsupervised} &34.6 & 33.6 &45.9 & 30.1  & 32.0 & 41.7 & 44.5 & 32.6  \\
			CDS  & \textbf{59.5} &   \textbf{62.8} & \textbf{71.5}  & \textbf{68.3} & \textbf{48.7} & \textbf{52.2} & \textbf{53.6} &  \textbf{52.4}    \\
			\bottomrule[1.0pt]

	\end{tabular}}
	\vspace{0.1em}
	\caption{Comparison with self-supervised learning baselines.
	}
	\label{exp:comparison_selfsup}
	\vspace{-2.5em}
\end{table}

\smallskip

In the column of \emph{Domain Adaptation}, we measure the target accuracy using CDAN and MME with different pre-training methods on  the Office 1-shot setting and Office-Home 3\% labels setting.
CDS significantly outperforms these baselines by a large margin in all cases. These results show that CDS can learn discriminative and domain-invariant features on both domains. 
To see where this performance gain comes from, we conduct \emph{Feature Analysis} in Table~\ref{exp:comparison_selfsup}.
Following the standard protocol suggested by Wu~\etal~\cite{wu2018unsupervised}, given the learned features (\ie, fixed features) from each method, we measure the target accuracy of a linear classifier trained on source features and source labels, and similarly, we measure the target accuracy using weighted $k$-nearest neighbors with source features and source labels. 
This directly assesses the quality of the learned representation. The results evidently show that the performance gain of the domain adaptation test mainly comes from the quality of the representation as shown by large performance margins in \emph{Feature Analysis}.

\smallskip
\subsection{Additional Analyses}

\label{sec:detail_analysis}

\begin{figure}[t]
	\centering
	\includegraphics[width=\linewidth]{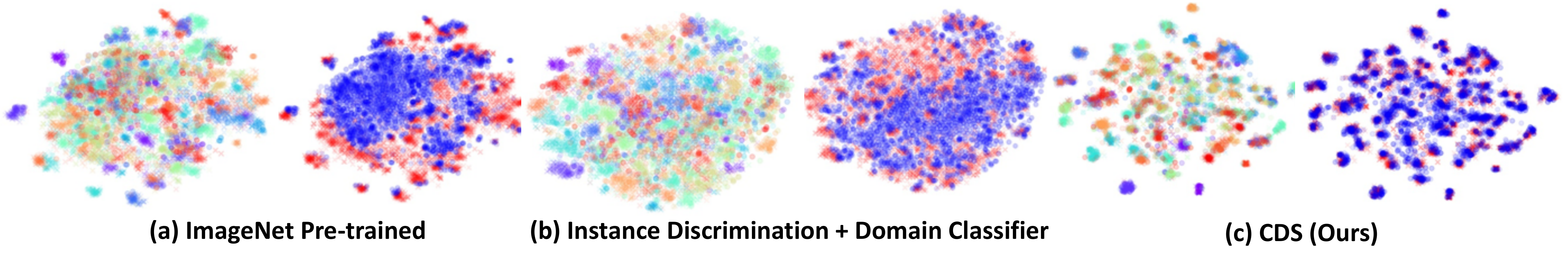}\vspace{-3mm}
	\caption{t-SNE visualization of ours and baselines. Compared to the others, CDS extracts features that are clearly class-discriminative as well as domain-invariant.
	We use two color coding schemes to represent different classes (left subfigures) and source-target (red-blue) domains (right subfigures), respectively.
	}
	\label{exp:tsne}
	\vspace{-2.0mm}
\end{figure}

\vspace{2mm}\noindent
\textbf{Does our method really learn class-discriminative and domain invariant feature?}\quad
Figure~\ref{exp:tsne} shows t-SNE visualization~\cite{maaten2008visualizing} of features obtained from the ImageNet pre-training and ours on the 
Real-to-Clipart setting in Office-Home. Compared to the adversarial domain classifier (DC) for feature alignment~\cite{ganin2017domain}, it qualitatively shows that CDS clusters the same class examples in the feature space; thus, CDS favors more discriminative features. In the figure, the red-blue dot plots represent the source and target domain data formation, which clearly evidences CDS generates well-aligned and domain-invariant features while preserving the class-discriminative power.

\begin{figure}[t]
	\centering
	\includegraphics[width=\linewidth]{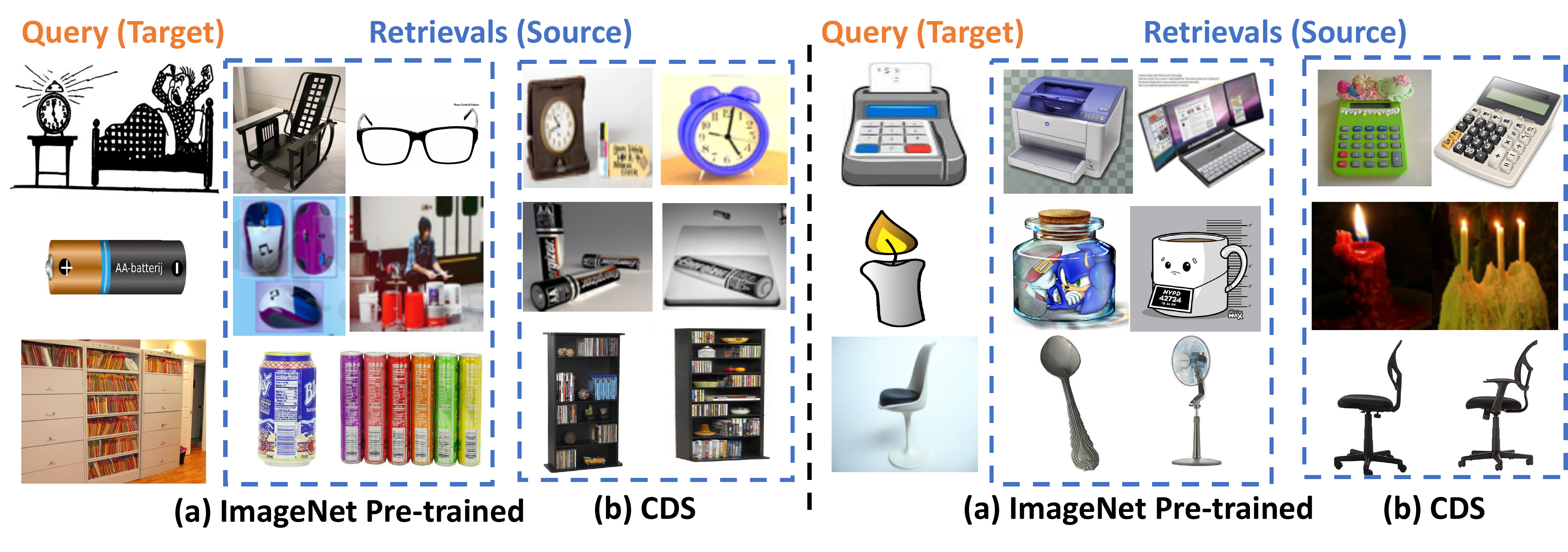}\vspace{-3mm}
	\caption{\small Retrieval of the closest cross-domain neighbors using standard ImageNet-pretrained features (a) and CDS (b). While ImageNet-pretrained features are biased to some textures and colors, our method learns semantic similarity between domains.
	}
    \label{fig:fig_qual_2}
    \vspace{-2mm}
\end{figure}

\vspace{2mm}\noindent
\textbf{What aspect of the feature is enhanced by our method?}\quad
Figure~\ref{fig:fig1}-(b,c) and Figure~\ref{fig:fig_qual_2}-(a,b) compare the retrieval results when using the network weights pretrained by ImageNet and ours on Office-Home. We observe that the weights pre-trained by ImageNet are sensitive to biased color and texture information, whereas our pretrained weights tend to capture better shape representation with a proper balance of color and texture information.

\begin{figure}[t]
\centering
	\includegraphics[width=11.5cm]{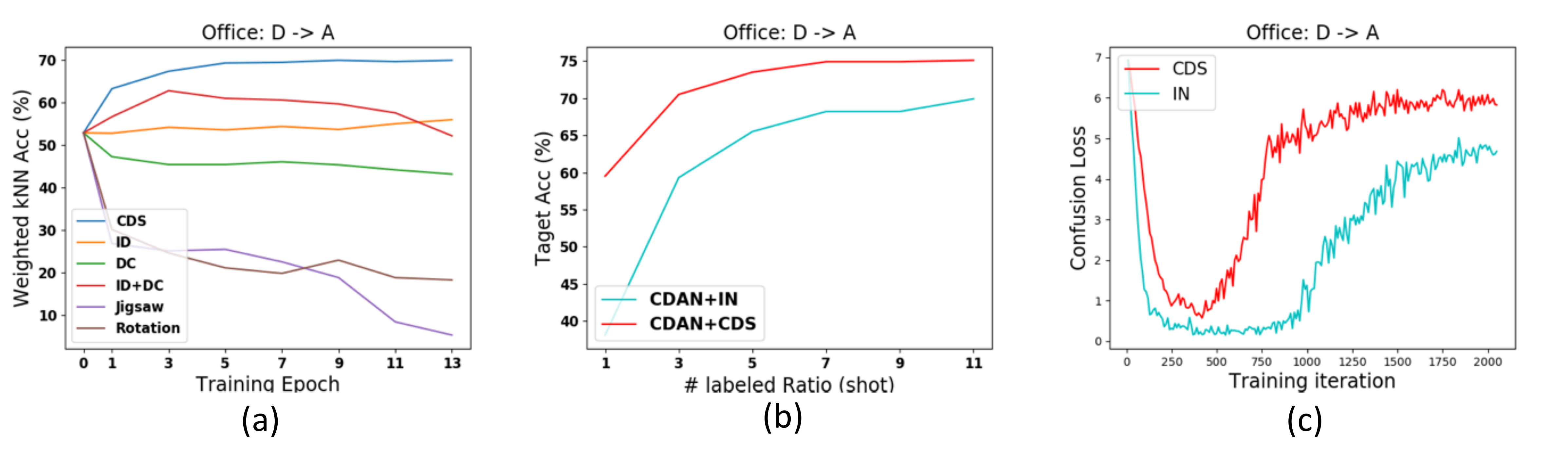}
	\vspace{-4mm}
	\caption{(a): Target accuracy using Weighted kNN according to training epochs. (b): Target accuracy according to the different number of labeled source examples. (c): Confusion loss (measured with the domain classifier) when using the pre-trained weights obtained by ImageNet (IN) and ours (CDS). }
	\label{exp:knn}
\end{figure}



\vspace{2mm}\noindent
\textbf{Assessment of Feature Quality according to Various Effects.}
We show the behavior according to the training epoch, the sample efficiency, and the domain gap on the feature space of our method. In Fig.~\ref{exp:knn}-(a), we measure the target accuracy using weighted k-nearest neighbors according to training epochs on the dslr-to-amazon setting on Office. The accuracy at epoch 0 reports the accuracy of ImageNet pre-trained weights. 
Note that Jigsaw and Rotation self-supervised methods decrease in accuracy over training; \ie, overfitting to the the respective proxy tasks. 
CDS consistently improves the performance across every training epochs, while even ID+DC suffers from overfitting from an intermediate step.
Also, Fig.~\ref{exp:knn}-(b) shows the sample efficiency to learn effective feature.
Our CDS consistently outperforms the competing baseline and stably improves the accuracy as the number of source labels is increased; Our CDS has a favorable sample efficiency.  
In Fig.~\ref{exp:knn}-(c), we show the target accuracy and confusion loss from the domain classifier used in the CDAN method according to training iterations. 
The confusion loss indicates
how the source and target features are aligned with each other. CDS obtains a higher confusion loss than that of the ImageNet pre-trained weights, which is another evidence that shows our features are more domain-invariant.


\subsection{Traditional Domain Adaptation with Full Source Labels}
\smallskip
\label{sec:traditional_domain_adaptation}
We also apply CDS to the traditional domain adaptation setting with full source labels. Table~\ref{exp:full_sup} shows the results on Office-Home and VisDA. CDS improves the performance of the state-of-the-art domain adaptation methods including MDD~\cite{zhang2019bridging}. This could be due to the features that are more discriminative on the target domain and domain aligned by our pre-trained network; and thereby it allows a model to transfer knowledge more easily for domain adaptation.



\begin{SCtable}[][t]
\centering
\resizebox{0.6\textwidth}{!}{\begin{tabular}{lr@{\hspace{3mm}}|@{\hspace{3mm}}lr@{\hspace{3mm}}|@{\hspace{3mm}}lr@{\hspace{3mm}}|@{\hspace{3mm}}lr@{\hspace{3mm}}|@{\hspace{3mm}}lr@{\hspace{3mm}}|@{\hspace{3mm}}lr}
\toprule[1.0pt]
\multicolumn{6}{c|@{\hspace{3mm}}}{Office-Home} &\multicolumn{6}{c}{VisDA } \\
\hline
\multicolumn{2}{c|@{\hspace{3mm}}}{DANN} & \multicolumn{2}{c|@{\hspace{3mm}}}{MME} & \multicolumn{2}{c|@{\hspace{3mm}}}{MDD} &\multicolumn{2}{c|@{\hspace{3mm}}}{DANN} & \multicolumn{2}{c|@{\hspace{3mm}}}{MME} & \multicolumn{2}{c@{\hspace{3mm}}}{CDAN} \\\hline
IN & 58.9 & IN & 66.4 & IN*& 67.3 & IN & 58.0 &IN & 68.9 & IN & 70.0 \\
CDS & \textbf{64.6} & CDS & \textbf{69.3} & CDS & \textbf{68.9} & CDS & \textbf{64.1} & CDS &\textbf{72.5} & CDS & \textbf{73.8} \\

\bottomrule[1.0pt]
\end{tabular}}
\caption{Target accuracy (AVG \%) on all settings in the traditional domain adaption with full source supervision. * denotes our reproduce of ~\cite{zhang2019bridging}.\vspace{-1.0em}}
\label{exp:full_sup}
\end{SCtable}


	\section{Conclusions}
Traditional domain adaptation assumes many and fully labeled source domain and only considers transfering knowledge. In this work, we investigate a new domain adaptation task which is more practical and challenging, where there are only few source labels available and many unlabeled source data. To leverage many unlabeled data and consider domain gap in different domains, we propose a novel Cross-Domain Self-supervised learning (CDS) which learns discriminative and domain-invariant features for domain adaptation. The pre-trained weights from CDS can be easily applied to boost performance of any domain adaptation method. We demonstrate that our method is also effective in the classical domain adaptation setting with the fully labeled source domain. To our knowledge, this work is the first to provide better pre-trained weights against ImageNet pre-trained weights which has been generally used in domain adaptation. 
		
	\bibliographystyle{splncs04}
	\bibliography{egbib}
	\newpage

\appendix
\subsubsection{Appendix.}
 We provide additional and detailed experiments which we cannot include in the main paper due to the space limit.
\smallskip
\begin{figure}[h]
\vspace{-2.0em}
    \centering
	\includegraphics[width=0.85\linewidth]{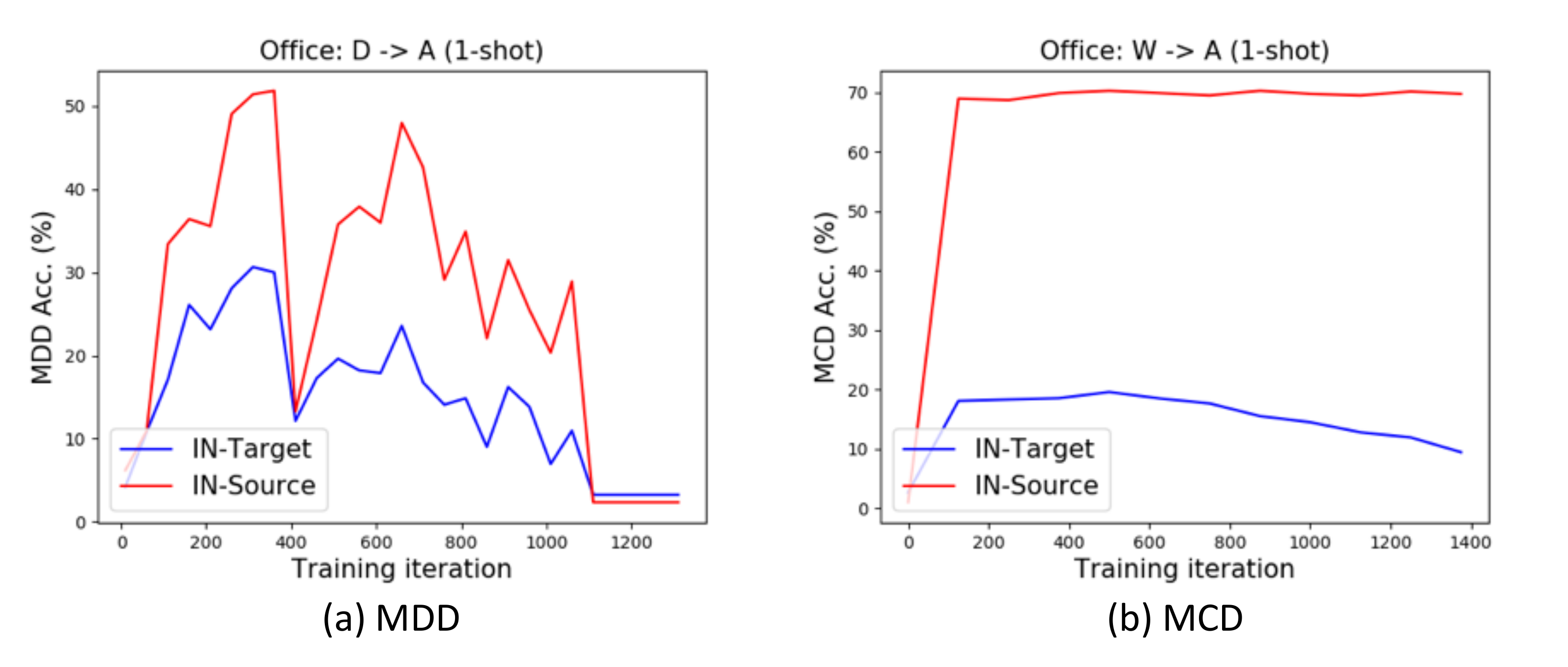}
	\caption{With few source labels, MDD~\cite{zhang2019bridging} and MCD~\cite{saito2018maximum} tend to have decreasing target accuracy with ImageNet pre-trained weights. (a): Target and source accuracy on MDD,  (b) Target and source accuracy on MCD.}
	\label{sup:sup_1}
	\vspace{-1.0em}
\end{figure}

\section{Detailed Analysis on MCD~\cite{saito2018maximum} and MDD~\cite{zhang2019bridging} with Few Source Labels}

From L446 to L449 of Sec 4.2 in the main paper, we also explore maximum classifier discrepancy based methods, MCD~\cite{saito2018maximum} and MDD~\cite{zhang2019bridging}. These methods minimize discrepancy of the task-specific decision boundaries or features from adversarially learned classifiers. We observe that MCD and MDD tend to have decreasing target accuracy during training in Office 1-shot settings. Figures~\ref{sup:sup_1}-(a,b) show the further analysis of these methods trained with few source labeled samples and unlabeled target samples. As shown in Fig.~\ref{sup:sup_1}-(a,b), we measure the target and source accuracy during training. For MDD, we observe that both source and target accuracy in  Fig.~\ref{sup:sup_1}-(a) are decreasing quickly and collapse to random predictions. We found that the adversarial learning on the source domain decreases the source accuracy. For MCD, we observe that the target accuracy is decreasing more slowly than that of MDD, while maintaining the source accuracy as shown in  Fig.~\ref{sup:sup_1}-(b). These results show that MCD and MDD need more labeled source examples to avoid collapsing to random predictions. 

\begin{figure}[h]
    \centering
	\includegraphics[width=0.85\linewidth]{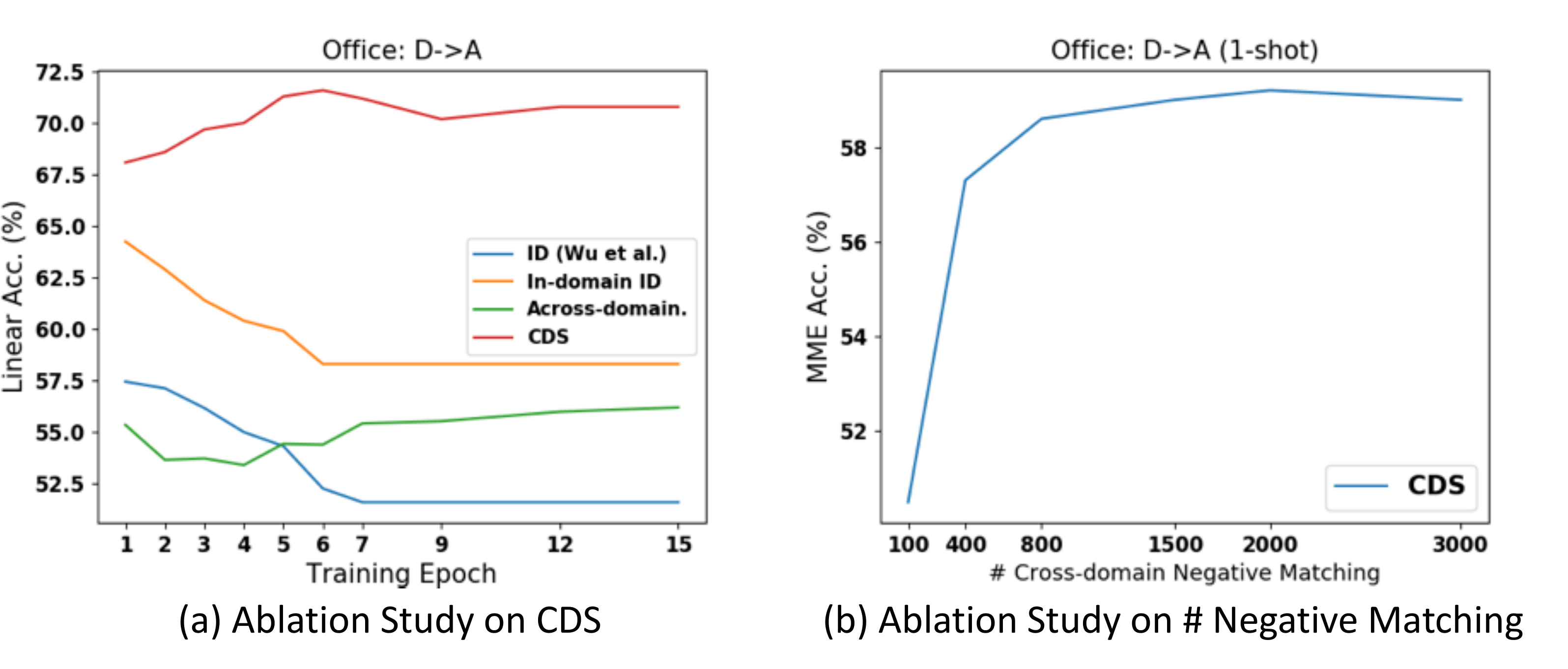}
	\caption{ (a): Ablation study on each component in CDS. Combining in-domain ID and cross-domain matching is necessary to improve accuracy. (b): Comparison of accuracy according to the number of negative matching pairs in cross-domain matching. It is important to discover negative matching as well as positive matching in cross-domain matching.}
	\label{sup:sup_2}
	\vspace{-1.0em}
\end{figure}

\section{Ablation Study on Each Component in CDS}
In Fig.~\ref{sup:sup_2}-(a), we perform an additional ablation study on our cross-domain self-supervised learning (CDS). We report comparisons with (1) Directly applying Instance Discrimination~\cite{wu2018unsupervised} (denoted by ID (Wu \etal), (2) In-domain ID (Sec 3.1) and (3) CDS (In-domain ID + cross-domain matching). We measure the target accuracy of a linear classifier trained on  learned source features (\ie, fixed features) and source labels from each method. By comparing the orange line and the blue line in Fig.~\ref{sup:sup_2}-(a), we see the increase in accuracy by performing in-domain ID rather than directly applying ID~\cite{wu2018unsupervised} to a domain adaptation task.  By comparing the green line (cross-domain matching) and the red line (CDS), it is important to combine both in-domain ID and cross-domain matching to increase accuracy.

\section{Importance of Discovering Cross-Domain Negative Matching Pairs}
In this section we show the importance of cross-domain negative matching as well as positive matching to ensure discriminative features when performing cross-domain feature matching (see the subsection ``Across-domain Self-supervision'' in the main paper for details). In Fig.~\ref{sup:sup_2}-(b), we compare adaptation accuracy by varying the number of negative matching pairs in cross-domain matching.  We measure the target accuracy of MME on the dslr-to-amazon 1-shot setting with few labeled source samples and unlabeled target samples. We consider top-K cross-domain nearest neighbors as negative matching. The x-axis represents the value of K and the y-axis represents the corresponding accuracy of MME. From  Fig.~\ref{sup:sup_2}-(b), we see that increasing number of cross-domain negative matching increases adaptation accuracy.

\begin{figure}[t]
\centering
\includegraphics[width=0.85\textwidth]{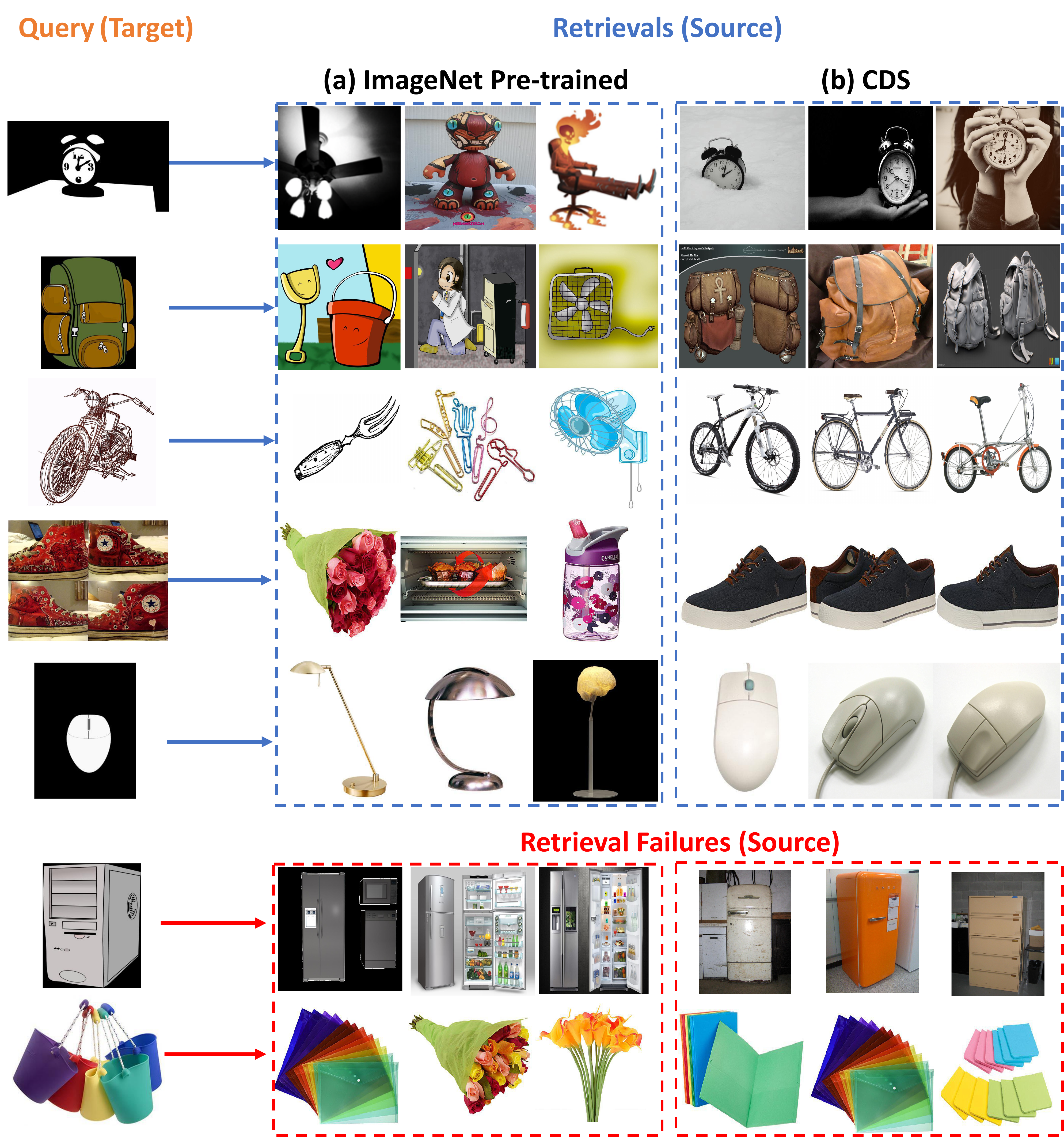}
\caption{Additional retrieval examples of the closest cross-domain neighbors using standard ImageNet pre-trained features (a) and CDS (b). The top section contained by the blue box shows the successful retrievals of semantic neighbors from CDS but the ImageNet pre-trained weights fail to retrieve semantic neighbors.  The bottom section contained with a red box shows some examples of failures of both methods.}
\label{fig:supp_quals}
\end{figure}

\section{Additional Retrieval Results}
We present additional retrieval results of the nearest cross-domain neighbors in Fig.~\ref{fig:supp_quals}. We measure the pairwise cosine similarity between a query feature in the target domain and features in the source domain. The features from the ImageNet pre-trained model are biased to some wrong texture information and visual clues, so that it does not provide semantically similar features for the same class images in different domains. We see that our method extracts semantically more meaningful features and provides discriminative features across domains. We also show the hard cases where our method also fails in the bottom section contained by the red box.

\section{Additional Results with Semi-supervised Learning Methods}
We provide the full results with the semi-supervised learning methods in addition to Table 5 in the main paper.  Please check Sec 4.2 for discussion.
\begin{table}[t]
 \resizebox{\textwidth}{!}{	
\begin{tabular}{rlc|c|c|c|c|c|c|c}
	\toprule[1.0pt]
		& & & \multicolumn{6}{c}{Office: Target Acc. (\%) / Source Acc. (\%) on 1-shot} \\
		\cline{4-10}
		& Adapt & Semi-sup. & A$\rightarrow$D & A$\rightarrow$W & D$\rightarrow$ A & D$\rightarrow$W & W$\rightarrow$A & W$\rightarrow$D & AVG\\
		\hline
		\hline
		&  \multicolumn{9}{c}{ (a) ImageNet pre-trained}\\
		\hline
		&SO & - & 28.3 / 39.8 & 31.6 / 42.0 & 34.8 / 71.3 & 64.5 / 68.5 & 37.0 / 64.0 & 56.8 / 64.1 & 42.2 / 58.3\\[0.5mm]
		&SO & ENT & 30.5 / 41.7 & 25.9 / 35.4 & 35.9 / 72.0 & 67.9 / 70.2 & 36.4 / 61.0 & 49.4 / 58.4 & 41.0 / 56.5 \\[0.5mm]
		&SO & ENT+VAT & 27.9 / 38.4 & 31.5 / 39.3 & 37.7 / 69.6 & 69.9 / 71.1 & 36.5 / 62.7 & 48.6 / 59.2 & 42.0 / 56.7\\
		\hline
		&CDAN~\cite{long2018conditional} & - & 32.5 / 41.1 & 29.2 / 34.6 & 38.1 / 67.9 & 70.7 / 69.6 & 34.8 / 56.2 & 64.1 / 65.7 & 44.9 / 55.9\\
		&CDAN & ENT & 31.5 / 42.0 & 26.4 / 35.4 & 39.1 / 72.4 & 70.4 / 70.7 & 37.5 / 59.8 & 61.9 / 61.4 & 44.5 / 57.0\\
		&CDAN&ENT+VAT & 32.7 / 43.1 & 34.6 / 39.7 & 39.8 / 68.5 & 69.0 / 70.2 & 38.6 / 63.2 & 62.9 / 65.3 & 46.3 / 58.3 \\
		\hline
		&	MME~\cite{saito2019semi} & -& 50.4 / 36.3 & 49.6 / 34.5 & 47.2 / 67.9 & 78.2 / 77.9 & 46.1 / 66.9 & 66.9 / 67.9 & 56.4 /  58.7\\[0.5mm]
		&MME&ENT & 37.6 / 45.6 & 42.5 / 43.4 & 48.6 / 77.7 & 73.5 / 76.7 & 47.2 / 69.8 & 62.4 / 67.9 & 52.0 / 63.5\\[0.5mm]
		&MME&ENT+VAT & 40.1 / 47.6 & 41.6 / 42.2 & 48.1 / 77.7 & 63.7 / 76.7 & 46.8 / 70.4 & 62.2 / 67.9 & 50.4 / 63.8\\
		\hline
		\hline
		&  \multicolumn{9}{c}{ (b) \textbf{CDS (Ours)}}\\\hline
		&SO &- & 52.2 / 56.9 & 54.6 / 54.3 & 51.3 / 78.6 & 78.5 / 79.9 & 55.3 / 73.3 & 71.5 / 72.9 & 60.6 /  69.3 \\[0.5mm]
		&SO&ENT  & 53.8 / 57.9 & 53.8  / 57.4 & 55.9 / 81.4 & 81.1 / 81.8 & 56.4 / 76.7 & 72.5 / 74.7 & 62.3 / 71.7 \\[0.5mm]
		&SO&ENT+VAT & 53.4 / 57.9 & 56.7 / 56.9 & 56.2 / 80.7 & 80.8 / 80.7 & 56.2 / 76.6 & 73.5 / 74.9 & 62.8 / 71.3 \\
		\hline
		&CDAN & - & 52.6 / 57.4 & 59.8 / 54.2 & 59.5 / 72.0 & 81.1 / 79.2 & 57.5 / 75.7 & 83.7 / 81.4 & 65.7  / 70.0\\
		&CDAN&ENT  & 53.8 / 57.2 & 65.7 / 60.9 & 62.0 / 77.9 & 83.0 / 81.6 & 57.4 / 76.8 & 77.1 / 76.7 & 66.5 / 71.9\\
		&CDAN&ENT+VAT  & 54.8 / 57.5 & 64.7 / 61.1  & 62.0 / 77.9 & 81.6 / 81.8 & 56.7 / 76.7 & 78.9 / 77.4 & 66.5 / 73.7\\
		\hline
		&MME &- & 51.2 / 35.1 & 56.9 / 39.3 & 58.0 / 82.2 & 80.3 / 80.3 & 58.6 / 76.7 & 70.3 / 72.5 & 62.6 / 64.4 \\[0.5mm]
		&MME&ENT  & 54.4 / 58.6 & 57.2 / 56.4 & 62.8 / 82.7 & 83.3 / 85.2 & 62.6 / 76.4 & 77.1 / 80.8 & 66.2 / 73.3 \\[0.5mm]
		&MME&ENT+VAT  & 53.2 / 58.5 & 58.1 / 54.1 & 69.5 / 88.1 & 82.5 / 85.0 & 62.2 / 76.6 & 76.9 / 80.9 & 67.1 / 73.9\\
		\bottomrule[1.0pt]
\end{tabular}}
\vspace{0.1em}
\caption{Additional results with semi-supervised learning methods in supplements to Table 5 in the main paper. Target and source accuracy (\%)  on unlabeled samples with semi-supervised learning methods.}
\label{tab:office_1_shot_semi}
\end{table}

\section{Adaptation Results with Multiple Runs} In order to show the stability of CDS, we perform multiple runs with three different random seeds. Table~\ref{exp:multi_run} reports the averaged accuracy and standard deviation of the three runs on the dslr-to-amazon 1-shot and 3-shots setting. The ImageNet pre-trained network obtains larger standard deviations on 1-shot compared to that of 3-shots. 

\begin{SCtable}[][t]
\centering
\resizebox{0.5\textwidth}{!}{	\begin{tabular}{c||c|c|c}
	\toprule[1.0pt]
		Adapt& Pretrain & 1-shot D$\rightarrow$A& 3-shots D$\rightarrow$A \\
		\cline{3-4}
		\hline
		\multirow{2}{*}{SO} &IN & 38.5 $\pm$ 1.8 & 51.8 $\pm$ 0.4\\
		&CDS & \textbf{54.8 $\pm$ 1.0} & \textbf{62.4 $\pm$ 0.9}\\
		\hline
		\multirow{2}{*}{CDAN} &IN & 42.1 $\pm$ 2.6 & 56.4 $\pm$ 2.1\\
		&CDS &  \textbf{60.2 $\pm $ \textbf{0.6}} & \textbf{71.5 $\pm$ 0.7}\\
		\bottomrule[1.0pt]
\end{tabular}}
\caption{Averaged  accuracy and standard deviation of three runs  on 1-shot and 3-shots for three multiple on the Office dataset. }
\label{exp:multi_run}
\vspace{-2.0em}
\end{SCtable}

\section{Additional Implementation Details}
We provide additional implementation details.
We set the dimensionality of the feature vector as 512 (\ie, $\mathbf{f} \in \mathbb{R}^{512}$). In the pre-training stage, we use all the samples in the source and target domain. Based on the validation set, we choose $|B_s|=|B_t|=32$ for Office-Home and Office, and $|B_s|=64, |B_t|=32$ for VisDA.

\begin{table}[t] 
\resizebox{\textwidth}{!}{\begin{tabular}{c||c|c|c|c|c|c|c|c|c|c|c|c|c|c}
\toprule[1.0pt]
		\multirow{2}{*}{Adapt}	&	\multirow{2}{*}{Pretrain} & \multicolumn{13}{c}{Office-Home: Target Acc. (\%)} \\
		\cline{3-15}
	 & & Ar$\veryshortarrow$Cl &Ar$\veryshortarrow$Pr&Ar$\veryshortarrow$Rw&Cl$\veryshortarrow$Ar&Cl$\veryshortarrow$Pr&Cl$\veryshortarrow$Rw&Pr$\veryshortarrow$Ar&Pr$\veryshortarrow$Cl&Pr$\veryshortarrow$Rw&Rw$\veryshortarrow$Ar&Rw $\veryshortarrow$Cl&Rw$\veryshortarrow$Pr& AVG \\
		\hline
		\hline
		\multirow{2}{*}{DANN~\cite{ganin2017domain}} & IN & 45.3 &	63.4&	72.0&	49.0&	59.5&	62.1&	47.1&	45.6	&69.3&	63.1&	52.7&	77.5&	58.9	\\
		& CDS & \textbf{51.3}&	\textbf{68.2}&	\textbf{74.3}&	\textbf{59.0}&	\textbf{64.1}&	\textbf{68.4}&	\textbf{58.1}&	\textbf{50.8}&	\textbf{75.0}&	\textbf{68.4}&	\textbf{59.6}&	\textbf{78.6}&	\textbf{64.6}
		\\
		\hline
		 \multirow{2}{*}{MME~\cite{saito2019semi}} & IN &54.2&	70.8&	74.8&	58.5&	69.4&	67.2&	57.9&	55.5&	77.0&	70.7&	58.6&	82.2 &	66.4 \\
		&  CDS & \textbf{56.9}&	\textbf{73.3}&	\textbf{76.5}&	\textbf{62.8}&	\textbf{73.1}&	\textbf{71.1}&	\textbf{63.0}	& \textbf{57.9} &	\textbf{79.4}&	\textbf{72.5}&	\textbf{62.5}&	\textbf{83.0}	& \textbf{69.3}
		 \\
		 \hline
		 \multirow{2}{*}{MDD~\cite{zhang2019bridging}} & IN &54.6&	\textbf{72.8}&	\textbf{78.3}&	57.9&	70.2&	71.8&	58.5&	52.9&	77.9&	\textbf{72.7}&	58.1&	81.8 & 67.3 \\
		 & CDS &\textbf{ 56.3}& 72.7 &	78.1&	\textbf{61.4}&	\textbf{72.0}&\textbf{	73.7}&	\textbf{64.1}&	\textbf{53.4}&	\textbf{79.4}&	72.5&	\textbf{60.9}&	\textbf{82.5}&	\textbf{68.9} \\
		 \bottomrule[1.0pt]
\end{tabular}}
\vspace{0.1em}
\caption{Detailed results on Table 7 in the main paper. We report target  accuracy (\%)  on each setting in Office-Home with full source labels. IN denotes the ImageNet pre-trained weights and CDS denotes our cross-domain self-supervised learning.}
\label{exp:full_office_home_supp}
\vspace{-1.0em}
\end{table}

\section{Detailed Results on Traditional Domain Adaptation with Full Source Labels}

For DANN, we use the implementation of~\cite{long2018conditional} and use their same validation set. Table~\ref{exp:full_office_home_supp} reports the accuracy on all settings on Office-Home. In DANN~\cite{ganin2017domain} and MME~\cite{saito2019semi}, CDS outperforms all of the settings in Office-Home.  In MMD~\cite{zhang2019bridging}, our method tends to improve the accuracy more in the settings with a large domain gap (\eg, Ar$\veryshortarrow$Cl) than a small domain gap (\eg, Ar$\veryshortarrow$Pr). For the settings with relatively high accuracy (\eg, Ar$\veryshortarrow$Pr and Ar$\veryshortarrow$Rw), CDS obtains very similar accuracy as the ImageNet pre-trained weights.

\end{document}